\documentclass[10pt,journal,cspaper,compsoc]{IEEEtran}
\ifCLASSINFOpdf
\else
\fi
\usepackage{amssymb}
\usepackage{epsfig}
\usepackage{epstopdf}
\usepackage{subfigure}
\usepackage{amsmath}
\usepackage{booktabs}
\usepackage{tabularx}
\usepackage{times}
\usepackage{graphicx}
\usepackage{amssymb}
\usepackage{color}
\usepackage{cite}
\usepackage{psfig}
\usepackage{algorithm}
\usepackage{algorithmic}
\usepackage{graphicx}
\usepackage{cases}
\usepackage{multirow}

\usepackage{psfig}
\usepackage{algorithm}
\usepackage{algorithmic}
\usepackage{graphicx}
\usepackage{amsmath}
\usepackage{booktabs}
\usepackage{subfigure}
\usepackage{cases}
\usepackage{enumerate}
\usepackage{wrapfig}
\usepackage{picinpar}
\usepackage{pbox}
\usepackage{multicol}
\usepackage{multirow}
\usepackage{amsmath}
\usepackage{comment}
\newcommand{\tabincell}[2]{\begin{tabular}{@{}#1@{}}#2\end{tabular}}

\newcommand{\thickhline}{%
	\noalign {\ifnum 0=`}\fi \hrule height 1pt
	\futurelet \reserved@a \@xhline
}
\begin{document}
%


\title{Fine-grained Text and Image Guided Point Cloud Completion with CLIP Model}

\author{
\IEEEauthorblockN{Wei Song, Jun Zhou, Mingjie Wang, Hongchen Tan, Nannan Li, Xiuping Liu} \\
\thanks{W. Song, J. Zhou and N. Li are with the School of Information Science and Technology, Dalian Maritime University, Dalian, China (E-mail: songwei100110@gmail.com, zj.9004@gmail.com, nannanli@dlmu.edu.cn). }
\thanks{M. Wang is with the School of Science, Zhejiang Sci-Tech University, Zhe Jiang, China (E-mail: mingjiew@zstu.edu.cn). }
\thanks{H. Tan is with the Institute of Artificial Intelligence,  Beijing University of Technology, Beijing , China (E-mail: tanhongchenphd@bjut.edu.cn). }
\thanks{X. Liu is with the School of Mathematical Sciences, Dalian University of Technology, Dalian 116024, China. (E-mail: xpLiu@dlut.edu.cn). }
}

%



\IEEEtitleabstractindextext{%
\begin{abstract}
This paper focuses on the recently popular task of point cloud completion guided by multimodal information. Although existing methods have achieved excellent performance by fusing auxiliary images, there are still some deficiencies, including the poor generalization ability of the model and insufficient fine-grained semantic information for extracted features. In this work, we propose a novel multimodal fusion network for point cloud completion, which can simultaneously fuse visual and textual information to predict the semantic and geometric characteristics of incomplete shapes effectively. Specifically, to overcome the lack of prior information caused by the small-scale dataset, we employ a pre-trained vision-language model that is trained with a large amount of image-text pairs. Therefore, the textual and visual encoders of this large-scale model have stronger generalization ability. Then, we propose a multi-stage feature fusion strategy to fuse the textual and visual features into the backbone network progressively. Meanwhile, to further explore the effectiveness of fine-grained text descriptions for point cloud completion, we also build a text corpus with fine-grained descriptions, which can provide richer geometric details for 3D shapes. The rich text descriptions can be used for training and evaluating our network. Extensive quantitative and qualitative experiments demonstrate the superior performance of our method compared to state-of-the-art point cloud completion networks.
\end{abstract}

\begin{IEEEkeywords}
Multimodal fusion, text corpus, point cloud completion
\end{IEEEkeywords}}
\maketitle
\IEEEdisplaynontitleabstractindextext

%
\IEEEpeerreviewmaketitle

%
%
%
%
\section{Introduction}
\begin{figure*}[ht]
  \includegraphics[width=0.95\textwidth]{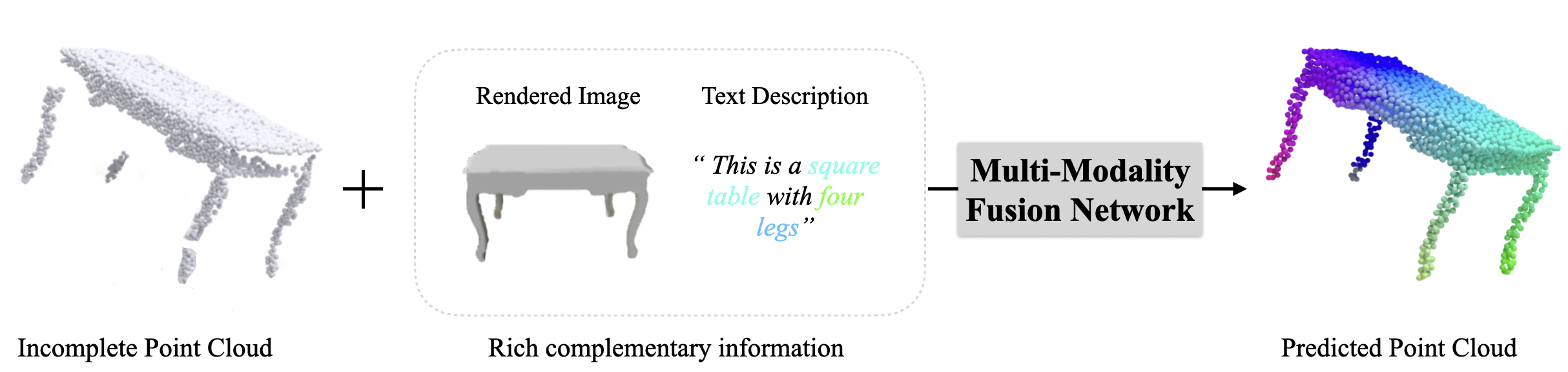}
  \caption{This is a novel point cloud completion method that can predict reliable complete shapes by leveraging the rich complementary information from a corresponding rendered image and rich text descriptions.}
  \label{fig:teaser}
\end{figure*}

\IEEEPARstart{W}{ith} the widespread use of 3D scanning devices such as LiDARs and RGB-D cameras, the acquisition of point clouds has become easier, which has promoted the development of a large number of research fields including automated machinery~\cite{pomerleau2015review}, driving autonomously~\cite{Kato_Tokunaga_Maruyama_Maeda_Hirabayashi_Kitsukawa_Monrroy_Ando_Fujii_Azumi_2018}, 3D visualization~\cite{wang2019applications}, scene understanding~\cite{hou20193d} and the production process~\cite{wells2021novel}. However, due to factors such as occlusion, reflection, transparency, low resolutions that may exist in the actual scanning process, the obtained point clouds are generally scattered and incomplete. The incomplete point clouds can produce ambiguous and misleading information, making it more difficult to recognize and understand 3D shapes. This can severely limit the performance of some downstream tasks such as 3D point cloud reconstruction~\cite{ma2018review,mandikal2019dense,choe2021deep}, detection and classification~\cite{grilli2017review, guo2020deep, fernandes2021point, li2021tutorial, mohammadi2021pointview, wang2022cagroupd}. Therefore, the research of point cloud completion has attracted increasing attention from the 3D visual community in recent years.

Although deep neural network-based methods have achieved excellent performance in point cloud completion, the lack of geometric cues and the partial sparseness of the scanned point cloud have hindered the further development of 3D point cloud completion task. Compared with the scanned point clouds, images can provide superior resolution and richer textures. Meanwhile, cross-sensors such as depth cameras and camera-LiDAR scanners~\cite{zhou2021tightly} are now available and less expensive to collect both colorful images and point clouds of 3D scenes, where images can provide more complementary geometry and semantic cues. The effectiveness of the image modality information introduction has been verified in recent literatures~\cite{zhang2021view, aiello2022cross, zhu2023csdn}. However, the existing methods mainly rely on the multi-view images and point clouds provided by the ShapeNet-ViPC dataset, which is a small-scale dataset. Insufficient training data leads to poor generalization ability of the trained network, making it difficult for the network to predict missing semantic and geometric details of incomplete shapes.

Furthermore, humans are considered to stand out among the animal world for their special ability to use language to describe objects, environments and events. Language can not only describe the physical and functional properties of objects, but also distinguish subtle differences between these properties through rich modifiers.
Recent advances in the fields of language modeling~\cite{Devlin_Chang_Lee_Toutanova_2019,brown2020language}, text-conditioned image generation~\cite{xu2018attngan,zhang2021cross,nichol2021glide,ramesh2021zero,ramesh2022hierarchical,ruiz2023dreambooth} and text-conditioned 3D shape generation~\cite{jain2022zero,wang2022clip,poole2022dreamfusion,lin2023magic3d,tsalicoglou2023textmesh, cao2023dreamavatar,richardson2023texture,zhang2023text2nerf} have demonstrated the effectiveness of textual descriptions as an efficient and valuable source of information. Therefore, text descriptions, as another important modality information, must be able to characterize richer shape semantics for point cloud completion task. However, due to the lack of a dataset including enough pairs of texts and point clouds, there remains an unexplored area in investigating the use of text as auxiliary information for point cloud completion. Although Text2Shape~\cite{chen2019text2shape} has exhibited promising results by curating a considerable dataset of natural language descriptions for various 3D objects, especially chairs and tables from the ShapeNet dataset~\cite{zhang2021view}. Regrettably, given the associated costs, this approach is difficult to apply to constructing larger-scale and multi-category 3D object text dataset. Furthermore, the text descriptions generated by Text2Shape~\cite{chen2019text2shape} cover multiple aspects of 3D objects, including color, texture and other shape details, but lacks sufficient geometric appearance information. As a result, this leads to redundant text descriptions that may exceed the input capacity of the CLIP~\cite{radford2021learning} text branch (77 tokens only), and more geometric details may be lost when extracting textual features.

In this paper, we propose a novel multimodal point cloud completion network that can simultaneously fuse the extra visual and textual information, as shown in Fig.~\ref{fig:teaser}. First, unlike the previous paradigm of multimodal point cloud completion for point-image pair fusion, we introduce additional text descriptions into the model to improve the model's ability to perceive shape semantics. Then, considering the relatively small scale of training data in the previous multimodal point cloud completion methods~\cite{zhang2021view, aiello2022cross, zhu2023csdn}, this leads to insufficient training of the extra modality branch. Therefore, inspired by the works~\cite{simonyan2014very, dosovitskiy2020image, radford2021learning, kirillov2023segment}, which introduce pre-trained models on large-scale dataset and demonstrate great generalization to a wide range of tasks, we transfer the pre-trained multimodal knowledge including visual and textual modalities from CLIP to our point cloud completion network. Compared with the scale of about 608k trained samples in the ShapeNet-ViPC dataset~\cite{zhang2021view}, the pre-trained CLIP model uses about 400 million training samples, ensuring that more effective textual and visual features can be extracted. Moreover, to address the limited scale and redundancy issues encountered with Text2Shape, and to mitigate the lack of text descriptions for 3D shapes in ShapeNet-ViPC~\cite{zhang2021view}, we introduce the ViPC-Text dataset. This dedicated dataset focuses on capturing descriptive information concerning geometric parts within 3D shapes. It is worth noting that we propose an innovative approach for generating large-scale and compact text descriptions, thereby facilitating the generation of text descriptions for arbitrary 3D shapes.

In summary, the contributions of this study are threefold:
\begin{itemize}
    \item Considering the generalization and reliability of large models, a pre-trained model is employed to effectively transfer multimodal features extracted from images and texts into our multimodal completion framework.
    \item A large-scale fine-grained corpus of 3D shapes named ViPC-Text is introduced to further explore the complementarity of text descriptions of 3D geometry with the missing semantics and structure of incomplete shapes. Furthermore, the method for generating 3D text descriptions can also be flexibly extended to arbitrary 3D shapes.
    \item A novel point cloud completion network is designed to simultaneously use three modality information including point clouds, images and texts. Extensive experiments demonstrate the superior performance of our method against previous state-of-the-art methods.
\end{itemize}

\label{secIntroduction}

\section{Related Work}
In recent years, stimulated by the great success of deep learning~\cite{LeCun_Bengio_Hinton_2015} in various vision tasks~\cite{guo2016deep}, the learning-based paradigms have become the dominant direction in the realm of point cloud completion~\cite{fei2022comprehensive} and has achieve outstanding performance. Existing learning-based point cloud completion can be classified into two types: unimodal and multimodal methods. In this section, we review these two types of methods in detail. Furthermore, we briefly introduce CLIP technology, which is employed in our proposed network for multimodal feature extraction.
    \subsection{Unimodal Point Cloud Completion}\label{subsec:uba}
    Unimodal point cloud completion only relies on information from a given incomplete point cloud to predict the missing part. Among them, PCN~\cite{yuan2018pcn} is a pioneering work that directly operates on incomplete point clouds by using a point cloud network for predicting complete shapes. Compared to traditional methods, the PCN~\cite{yuan2018pcn} avoids the need for any prior assumptions regarding the shape's structure such as symmetry, or requiring annotations related to the underlying shape such as category information. Besides, benefiting from the design of a folding-based encoder, this method is able to generate the missing details of the incomplete shapes while maintaining a relatively low number of parameters. Subsequently, TopNet~\cite{tchapmi2019topnet} designs a hierarchical decoder with the root tree structure, which can progressively generate a series of sub-nodes containing structured point clouds. Thus, the subset of points contained in the child nodes can be pieced together to form a final complete shape. MSN~\cite{liu2020morphing} is also a coarse-to-fine completion method, which can predict complete shapes by assembling a collection of estimated surface elements. Recently, inspired by transformer technology~\cite{vaswani2017attention} used in the visual tasks, the architectures of Point Transformer~\cite{Guo_Cai_Liu_Mu_Martin_Hu_2021, Pan_Xia_Song_Li_Huang_2021} are proposed and widely used in the tasks of point cloud processing. Based on this technology, PointTr~\cite{yu2021pointr} represents the point clouds as a set of unordered groups of points with position embeddings and converts the point clouds to a sequence of proxies under the way of encoder-decoder. SnowflakeNet~\cite{xiang2021snowflakenet} utilizes a Skip-Transformer to generate child points by gradually splitting parent points and predicting the complete shapes with more details. Also using the transformer technology, some recent works~\cite{zhou2022seedformer,Chen_Long_Qiu_Yao_Zhou_Luo_Mei,Li_Gao_Tan_Wei_Pointr} employ a coarse-to-fine approach based on proxy points for point cloud completion. Specifically, Seedformer~\cite{zhou2022seedformer} primarily utilizes seeded proxies to complete the point cloud through point upsampling layers. Then, ProxyFormer~\cite{Li_Gao_Tan_Wei_Pointr} divides the point clouds into existing and missing parts and facilitates communication between the two parts using proxies. Besides, AnchorFormer~\cite{Chen_Long_Qiu_Yao_Zhou_Luo_Mei} models regional discrimination by learning a set of anchors based on the point features of the input partial observation, which further employs a modulation scheme to transform a canonical 2D grid into a detailed 3D structure at specific locations of the sparse points. Although this kind of methods have achieved good performance, it is difficult for unimodal techniques to overcome the ambiguity of incomplete data itself, which makes it difficult for 3D completion methods to make greater breakthroughs.
    
    \subsection{Multimodal Point Cloud Completion}\label{subsec:mba}
    Multimodal completion task aims to utilize complementary modality information to improve the quality of the complete shapes, which is pioneered by ViPC~\cite{zhang2021view}. The motivation for this method is that the extra modalities contain the necessary structure or semantic information of the missing part of a given shape. Particularly, ViPC~\cite{zhang2021view} uses an image to predict the global shape information that is lacking in the incomplete point cloud. Then, the predicted coarse point cloud from the single-view image is used to guide the fine-grained completion. However, due to the lack of direct multimodal feature fusion at the coarse-grained stage, the use of complementary information in images is not sufficient. To solve this problem, XMFNet~\cite{aiello2022cross} proposes a multimodal feature fusion network to effectively combine features extracted from the two modalities in a localized latent space, thus avoiding the hard inverse problem of directly reconstructing point clouds from images. Moreover, CSDN~\cite{zhu2023csdn} proposes a cross-modal shape-transfer dual-refinement network, so that the auxiliary image can participate in the coarse-to-fine completion pipeline in the whole cycle. The guidance of multimodal information can significantly improve the quality of complete shapes, and our proposed method also adopts the paradigm of multimodal fusion. But unlike the methods mentioned above, we introduce a pre-trained large model to greatly increase the number of samples, which ensures that the multimodal complementary information can be effectively extracted. In addition, we expand the category of multimodal information by adding text descriptions to further increase complementary semantic information. Last but not least, a corpus with rich fine-grained geometric information is also proposed, and the text descriptions can effectively improve the semantic and geometric structure prediction capabilities of multimodal networks for incomplete shapes.
    
    \subsection{CLIP-based 3D Task}\label{subsec:clip}
    Recently, CLIP~\cite{radford2021learning} represents a noteworthy advancement in training visual models by natural language supervision, resulting in strong alignment between textual modal data and visual modal data. This alignment enables the models to uncover important feature information from novel perspectives that may be present in the other modality. Inspired by CLIP, CLIP-Based 3D point cloud analysis tasks have attracted increasing attention from the research community. For the 3D shape understanding task, PointCLIP~\cite{zhang2021pointclip} and PointCLIP V2~\cite{Zhu2022PointCLIPV2} convert 3D representation to multi-view image representation and attempt to process 3D information using CLIP~\cite{radford2021learning}, which can achieve knowledge transfer from 2D to 3D and conduct 3D recognition via CLIP pre-trained in 2D. However, these works do not directly use point cloud representation, resulting in insufficient geometric feature extraction. Differently, the current works ~\cite{zeng2023clip, hegde2023clip, xue2022ulip, xue2023ulip}  attempt to use natural language supervision to train the triplets including point clouds, corresponding rendered 2D images, and text descriptions, thus directly obtaining the feature representation of the point cloud. Therefore, the 3D features extracted from this kind of method can achieve zero-shot capabilities, which are similar to the features extracted from CLIP. The above methods have demonstrated the generalization and strong representation ability of CLIP for 3D tasks. Inspired by this, we also propose a CLIP-based multimodal guided point cloud completion network.

\label{secRelatedWork}

\section{method}

    \begin{figure*}[htp!]
      \centering
      \includegraphics[width=0.95\textwidth]{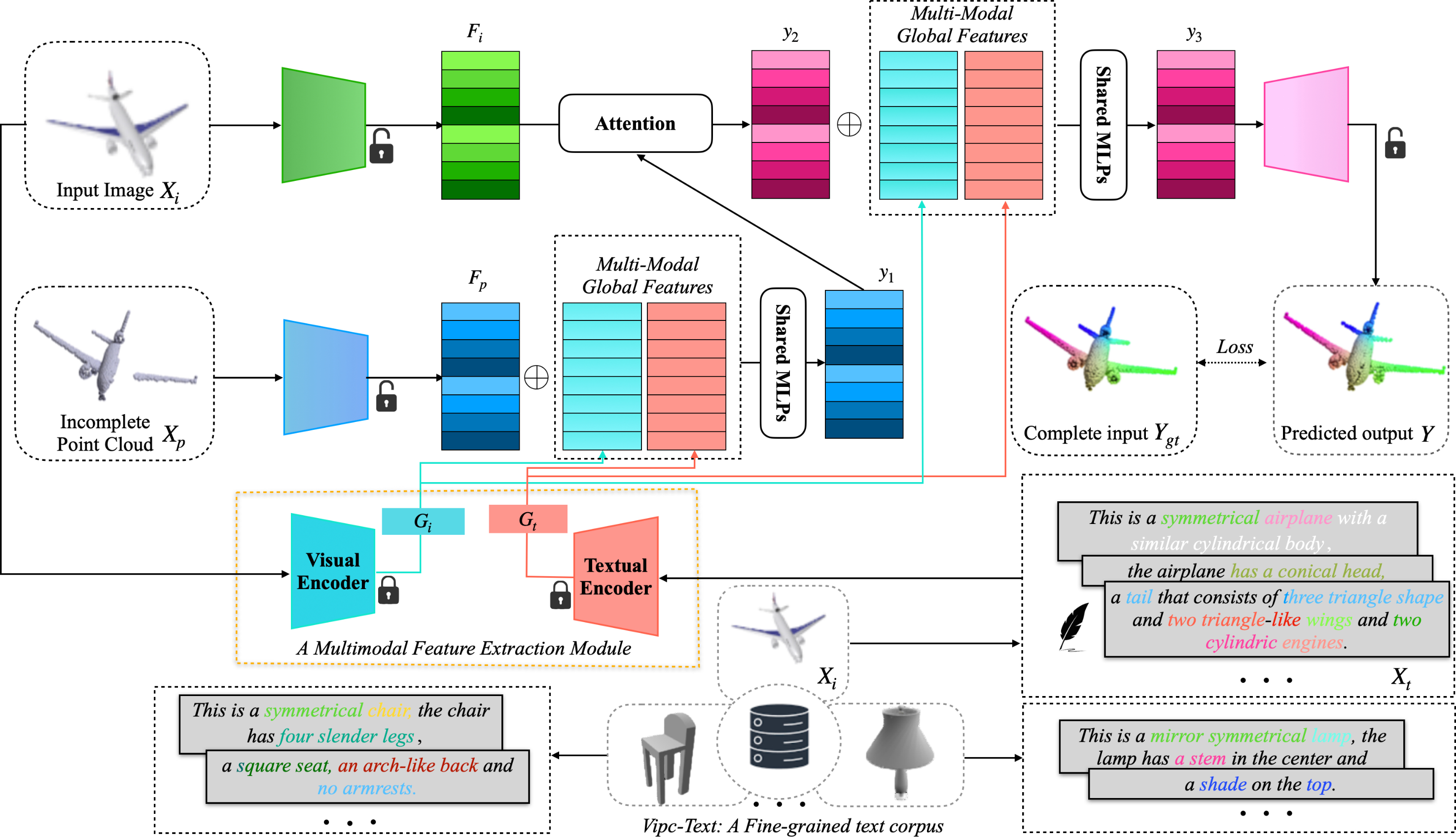}
    \caption{Architecture of the proposed Fine-grained Text and Image Guided Point Cloud Completion using CLIP, see Sec.~\ref{Overview} in the text for more details.}
    \label{fig:overview_ftpnet}
    \end{figure*}

    \subsection{Overview}\label{Overview}
    The goal of this paper is to explore the effectiveness of rich multimodal information for point cloud completion. The pipeline of our network is shown in Fig.~\ref{fig:overview_ftpnet}. First, in order to take advantage of the generalization ability of the large model, we use the CLIP model to extract the features of multimodal information. Second, a multi-stage feature fusion strategy is used in our backbone network to achieve the effective fusion of multimodal information. Finally, we propose an efficient text generation algorithm and build a corpus with fine-grained geometric descriptions to further improve the model's understanding of semantics and geometric structures. In Sec.~\ref{subsec:The Architecture of FTPNet}, we first give an overview of our proposed multi-modality fusion network, which can achieve state-of-the-art performance. Then in Sec.~\ref{subsubsec: visual encoder and text encoder}, we briefly review the pre-trained visual and textual encoders employed in our network. Subsequently, We show the LLM-assisted 3D component corpus in Sec.~\ref{subsec: llm-assisted 3d componet corpus}. In Sec.~\ref{subsec: a multi-stage fusion method}, we describe the our proposed fusion strategy in detail. To the end, we show the loss function used for training and evaluating in Sec.~\ref{subsec: loss_and_evaluation_function}.
    
    \subsection{Overview of Multi-modality Fusion Network}\label{subsec:The Architecture of FTPNet}
    Inspired by XMFNet~\cite{aiello2022cross}, we design a novel architecture for 3D point cloud completion, which adopts richer modality information and a multi-stage fusion strategy. Similar to XMFNet~\cite{aiello2022cross}, the basic network consists of two modality-specific feature extractors to construct the fine-grained features for point clouds and images, and then an attention module is employed for feature fusion, as shown in the upper part of Fig.~\ref{fig:overview_ftpnet}. Here, an incomplete point cloud $X_{p} \in \mathbb{R}^{N \times 3}$ and a randomly selected rendered images $X_{i} \in \mathbb{R}^{3 \times 224 \times 224}$ from 24 perspectives are fed into the network, then a complete point cloud $Y \in \mathbb{R}^{2048 \times 3}$ is predicted. Since this architecture only introduces the multimodal information from images and point clouds and only trains on a small-scale dataset, the semantic information acquired by this basic network is not sufficient and the generalization ability of this network is poor. To solve this problem, we introduce a pre-trained CLIP model~\cite{radford2021learning} on top of this basic network to extract more powerful multimodal information from images and text descriptions, as shown in the bottom of Fig.~\ref{fig:overview_ftpnet}. The pre-trained CLIP model has a more strong generalization ability and can obtain feature representations with more semantic information, which can help to improve the performance of the basic point cloud completion network. See the bottom of Fig.~\ref{fig:overview_ftpnet}, the pre-trained CLIP~\cite{radford2021learning} consists of textual and visual encoders, which can provide more comprehensive modal information to guide the completion process. Furthermore, to extract more information about the geometric features of the point cloud to improve the understanding of the point cloud, (see the bottom of Fig.~\ref{fig:overview_ftpnet}), we generate fine-grained geometric text descriptions for the point cloud, which can be used to accurately locate the semantic parts of the point cloud and provide rich geometric structure information. Ultimately, the global geometry-aware features $G_{i} \in R^{512 \times 1}$ and $G_{t} \in R^{512 \times 1}$ are available from the visual encoder and textual encoder respectively. Notably, we design a multi-stage multimodal feature fusion mechanism to ensure that multimodal features with richer semantic and geometric representations can effectively guide the process of point cloud completion. In the following, we present each module of the proposed method in more detail.

    \begin{figure*}[h] 
      \centering
      \includegraphics[width=\textwidth]{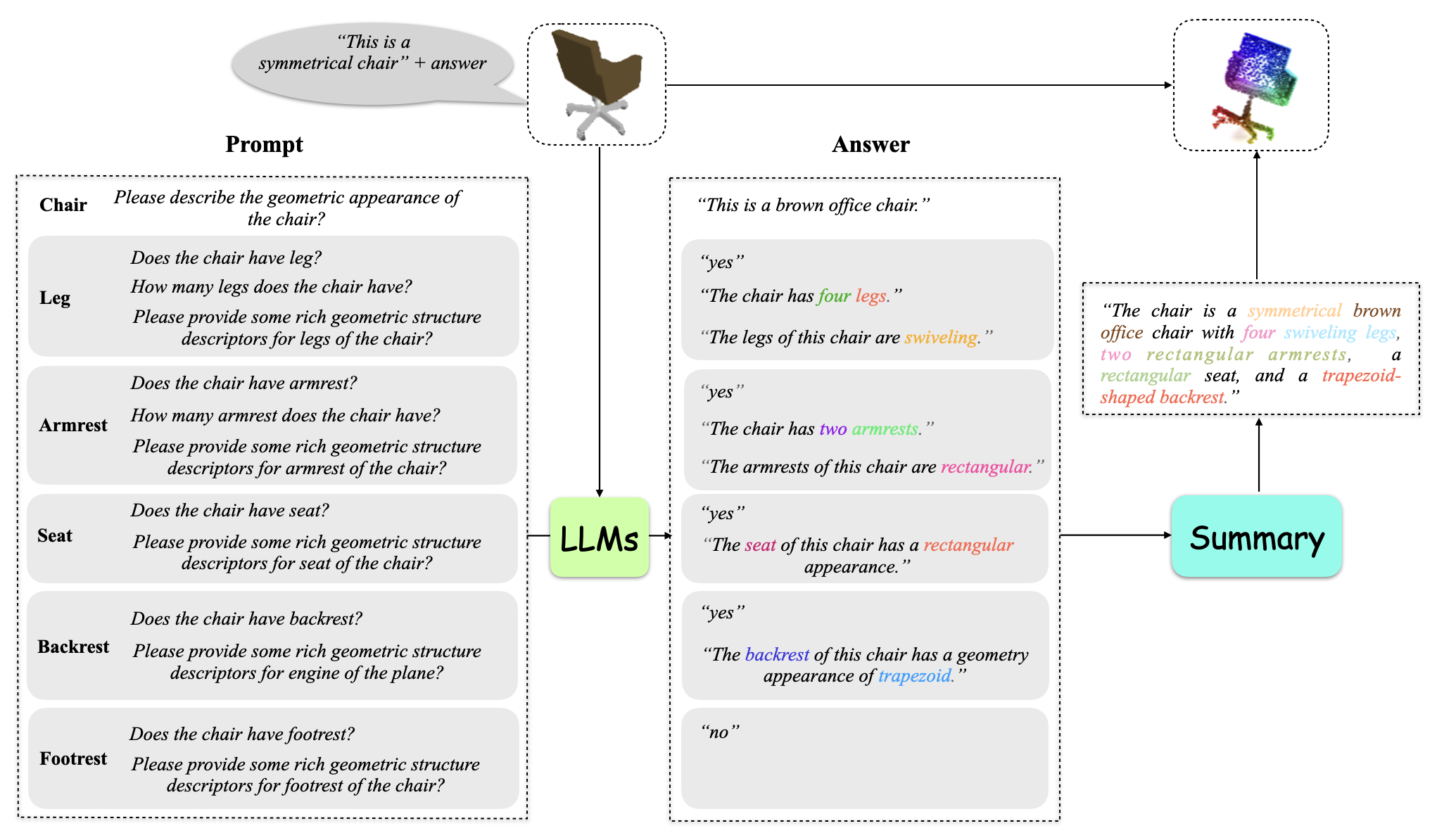}
      \caption{LLM-assisted 3D component corpus. We feed a query image and a series of part-related language commands into the pre-trained LLMs, which can automatically determine whether each component exists, and generate textual geometric descriptions for each component of the image.}
      \label{fig:text_llms}
    \end{figure*}
    
    \begin{figure*}[h] 
      \centering
      \includegraphics[width=0.95\textwidth]{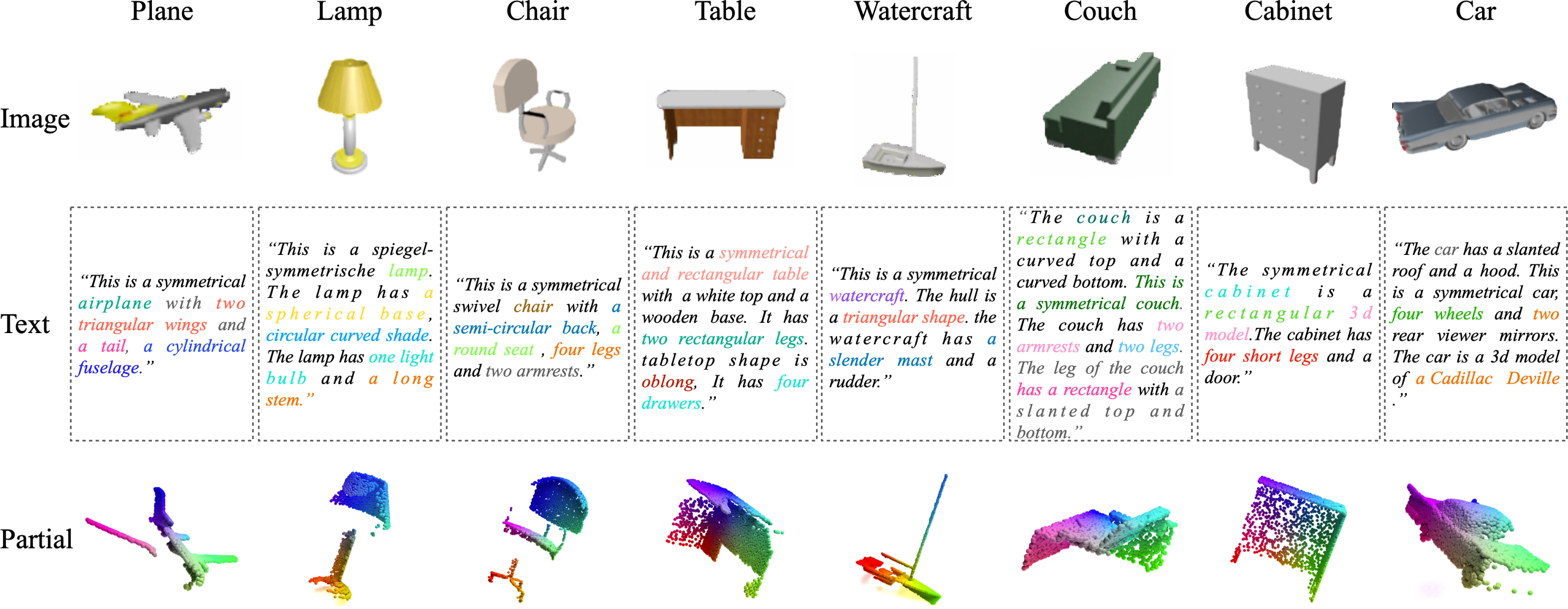}
      \caption{Example instances from our ViPC-Text dataset. The ViPC-Text dataset contains a large number of triples consisting of rendered images, fine-grained text descriptions and incomplete point clouds. The triple instances are obtained from a variety of categories. Among them, the text descriptions contain rich and fine-grained geometric descriptions of the 3D shapes.}
      \label{fig:example_3d}
    \end{figure*}

    \subsection{Visual and Textual Encoders}\label{subsubsec: visual encoder and text encoder}
    CLIP~\cite{radford2021learning} is a pre-trained model that includes a visual encoder and a textual encoder, which are learned by matching about 400 million image-text pairs. The scale of the training dataset ensures the generalization of the model and the alignment of multimodal features allows the two encoders to extract latent features with unified semantics and visual consistency, which are beneficial for the task of point cloud completion. In this study, we embed this pre-trained model into our completion architecture. First, rendered images derived from point clouds in the ShapeNet-ViPC dataset~\cite{zhang2021view} are utilized as input to the visual encoder of CLIP. Simultaneously, the textual encoder can accept a specific textual prompt that includes a defined template: “This is a [category]” and the “category” word should correspond to the shape category in the ShapeNet-ViPC dataset~\cite{zhang2021view}. Furthermore, a richer description such as “This is a classic styled, white and metal-based table” (replace with a better textual description containing structural information) can also follow the previous prompt as a semantic extension. The extra descriptions are formed as our proposed ViPC-Text dataset. Several examples of the corresponding point clouds, images and texts are displayed in Fig.~\ref{fig:example_3d}. The extracted latent features can be applied at multiple stages of the completion model to enhance the multimodal fusion. The subsequent quantitative experiments and visual comparison show the effectiveness of the extracted latent features.

    \subsection{LLM-assisted 3D Component Corpus} \label{subsec: llm-assisted 3d componet corpus}
    Obviously, text description, as a kind of modality information, can accurately describe the geometric appearance of 3D point clouds, and it is relatively easy to obtain a simple text description for a given 3D shape. It has been demonstrated in~\cite{zhang2021pointclip,Zhu2022PointCLIPV2,xue2022ulip, xue2023ulip} that it is feasible to use a text description to improve the understanding of point clouds. In these works, the text descriptions constructed to describe the 3D shape are simple and coarser-grained. Notably, incomplete shapes are more likely to be missing several components and the more accurate and fine-grained geometric descriptions are necessary for the task of point cloud completion. It is extremely difficult to construct a rich, accurate and detailed text corpus for a large number of 3D shapes, which requires a lot of manual annotations. Therefore, in this paper, we construct a fine-grained geometric text corpus for describing components of 3D shapes from the existing ShapeNet-ViPC dataset by using BLIP-2~\cite{Li_Li_Xiong_Hoi}, which belongs to large-scale language models (LLMs). It should be emphasized that the entire construction process is automatic and efficient.
    
    As shown in Fig.~\ref{fig:text_llms}, taking the chair model as an example, we show the pipeline of fine-grained geometric corpus generation. Generally, BLIP-2~\cite{Li_Li_Xiong_Hoi} receives an image and question of textual prompt and outputs an answer of textual prompt. In order to adapt BLIP-2~\cite{Li_Li_Xiong_Hoi} to describe the fine-grained geometric appearance of 3D shapes, we use the following series of language prompts:
    
    ~
    
    \noindent\textbf{Category Question Answering.} For a coarse-grained description of shape appearance, we construct a category-related question. e.g., Input: “Please describe the geometric appearance of the [chair]?”; Output: “This is a brown office [chair].”.
    
    ~

    \noindent\textbf{Existence Question Answering.} In order to judge whether a certain component exists in a given 3D shape, we designed an existence question, which can help us combine the output descriptions of multiple components efficiently.  e.g., Input: “Does the [chair] have [leg]?”; Output: “yes or no.”.

    ~

    \noindent\textbf{Quantity Question Answering.} In order to more accurately describe the quantity of a certain component for a given 3D shape, we design a question for the number of a certain component. e.g., Input: “How many [leg] does the [chair] have”; Output: “ The [chair] has four [legs].”.

    ~

    \noindent\textbf{Appearance Question Answering.} For a fine-grained description of component appearance, we construct a component-related question.  e.g., Input: “Please provide some rich geometric structure descriptors for [seat] of the [chair]?”; Output: “ The [seat] of this [chair] has a rectangular appearance”.

    ~

\begin{figure}[htb]
      \centering
      \includegraphics[width=\linewidth]{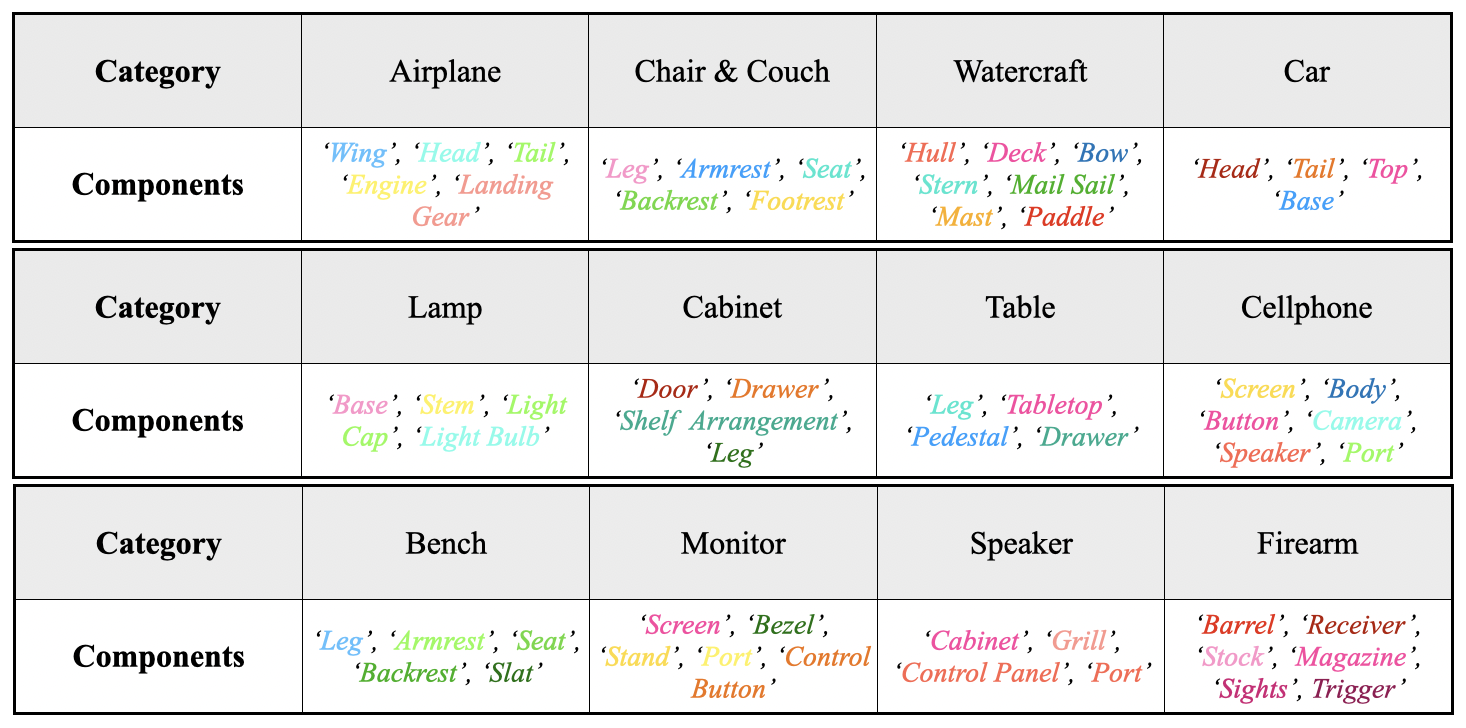}
      \caption{\label{fig:components}
            Here are components of each category in 13 categories from ShapeNet-ViPC.}
    \end{figure}

    \noindent Finally, we summarize the multiple outputs to form a fine-grained geometric appearance description of 3D shapes. In fact, due to the length of text descriptions generated by BLIP-2~\cite{Li_Li_Xiong_Hoi} always exceeds 258 words, we use the natural language denoising network Bart~\cite{lewis2019bart} to compress the length of text description to 50-58.
    In addition, in order to speed up the generation process of 3D prompts, we just select a random image under 24 views corresponding to each point cloud model as the input image of BLIP-2~\cite{Li_Li_Xiong_Hoi}. The output text description generated by BLIP-2~\cite{Li_Li_Xiong_Hoi} is then considered as the average text description overall 24 viewpoints for a given point cloud. Besides, we design different components for 13 categories utilized throughout the experiment (Fig.~\ref{fig:components} gives more details about the categories of the components).

    \subsection{A Multi-stage Fusion Strategy}\label{subsec: a multi-stage fusion method}
    Given the global features $G_{i}$ and $G_{t}$ that are respectively extracted from the visual and textual encoders of the multimodal feature extraction module, our aim is to fuse these multimodal features into our basic network to enhance the ability of the basic network to understand the structure and semantics of the incomplete 3D shapes. As shown in Fig.~\ref{fig:overview_ftpnet}, we perform the strategy of two-stage multimodal feature fusion in the basic network, where the first fusion operation is performed before fine-grained cross-attention fusion and the second one is performed after the attention fusion.

    Specifically, at the first stage of multimodal feature fusion, we concatenate the textual features $G_{i}$  and visual features $G_{i}$ together. Then, we repeat these multimodal features and concatenate the features with fine-grained point cloud features $F_{p} \in R^{256 \times 128}$. Subsequently, the operation of shared $MLPs$ is performed to achieve the fusion of multimodal features and the output can be denoted as $y_{1} \in R^{256 \times 128}$. After the first multimodal fusion, we fuse the fine-grained image features $F_{i}$ with the multimodal features $y_{1}$. Here, the operation of cross-attention is used to achieve the fine-grained feature fusion, and the output is denoted as $y_{2} \in R^{256 \times 128}$. Then, we use the fine-grained feature $y_{2}$ to perform the second fusion with multimodal features $G_{i}$ and $G_{t}$. Since the process of the second fusion is similar to the first one, the introduction is omitted here. The output by the second stage is denoted as $y_{3} \in R^{256 \times 128}$
     
    \subsection{Loss Function}\label{subsec: loss_and_evaluation_function}
    Similar to previous works~\cite{zhu2023csdn, zhang2021view}, we utilize the symmetric version of Chamfer Distance (CD) as our loss function, which can measure the difference between the predicted point cloud $Y$ and ground truth $Y_{gt}$. The loss function is written as: 
    \begin{equation}
    L_{cd} = \frac{1}{|Y|}\sum_{y \in Y}\min_{\hat{y} \in Y_{gt}}\|y-\hat{y}\|_{2}^{2} + \frac{1}{|Y_{gt}|}\sum_{\hat{y} \in Y_{gt}}\min_{y \in Y}\|\hat{y}-y\|_{2}^{2},
    \end{equation}
    where the first item pushes the predicted point cloud $Y$ as close to the ground truth $Y_{gt}$ as possible, and the second term enables $Y$ to cover the ground truth $Y_{gt}$. 
\label{secOverview}

\begin{table*}
    \tiny
    \renewcommand\arraystretch{1.2}
    \centering
    \caption{Quantitative comparisons to state-of-the-art methods on known eight categories of ShapeNet-ViPC dataset by using Mean Chamfer Distance per point ($\times10^{-3}$) with 2048 points. The best is highlighted in bold. $*$ means the code is not available or uncompleted.}
    \label{tab: cd of constrative experiments}
    \footnotesize
    \begin{tabular}{c|c|c|c|c|c|c|c|c|c}
    \hline
    \multirow{2}{*}{Methods} & \multicolumn{9}{c}{Mean CD per point (lower is better)} \cr\cline{2-10}
                                                   & Mean 
                                                   & Airplane 
                                                   & Cabinet 
                                                   & Car 
                                                   & Chair 
                                                   & Lamp 
                                                   & Sofa 
                                                   & table 
                                                   & Watercraft \cr
    \hline
    \hline
          \multicolumn{10}{c}{Unimodal Methods} \\
          \hline
          AtlasNet~\cite{groueix2018papier}        & 6.062 & 5.032 & 6.414 & 4.868 & 8.161 & 7.182 & 6.023 & 6.561 & 4.261 \\
          \hline
          FoldingNet~\cite{yang2018foldingnet}     & 6.271 & 5.242 & 6.958 & 5.307 & 8.823 & 6.504 & 6.368 & 7.080 & 3.882 \\
          \hline
          PCN~\cite{yuan2018pcn}                   & 5.619 & 4.246 & 6.409 & 4.840 & 7.441 & 6.331 & 5.668 & 6.508 & 3.510 \\
          \hline
          TopNet~\cite{tchapmi2019topnet}          & 4.976 & 3.710 & 5.629 & 4.530 & 6.391 & 5.547 & 5.281 & 5.381 & 3.35 \\
          \hline
          ECG~\cite{pan2020ecg}                    & 4.957 & 2.952 & 6.721 & 5.243 & 5.867 & 4.602 & .813 & 4.332 & 3.127\\
          \hline
          VRC-Net~\cite{pan2021variational}        & 4.598 & 2.813 & 6.108 & 4.932 & 5.342 & 4.103 & 6.614 & 3.953 & 2.925\\
          \hline
          PF-Net~\cite{huang2020pf}                & 3.873 & 2.515 & 4.453 & 3.602 & 4.478 & 5.185 & 4.113 & 3.838 & 2.871 \\
          \hline
          MSN~\cite{liu2020morphing}               & 3.793 & 2.038 & 5.06 & 4.322 & 4.135 & 4.247 & 4.183 & 3.976 & 2.379 \\
          \hline
          GRNet~\cite{xie2020grnet}                & 3.171 & 1.916 & 4.468 & 3.915 & 3.402 & 3.034 & 3.872 & 3.071 & 2.160 \\
          \hline
          PoinTr~\cite{yu2021pointr}               & 2.851 & 1.686 & 4.001 & 3.203 & 3.111 & 2.928 & 3.507 & 2.845 & 1.737 \\
          \hline
          PointAttN~\cite{wang2022pointattn}       & 2.853 & 1.613 & 3.969 & 3.257 & 3.157 & 3.058 & 3.406 & 2.787 & 1.872 \\
          \hline
          SDT~\cite{zhang2022point}                & 4.246 & 3.166 & 4.807 & 3.607 & 5.056 & 6.101 & 4.525 & 3.995 & 2.856 \\
          \hline
          Seedformer~\cite{zhou2022seedformer}     & 2.902 & 1.716 & 4.049 & 3.392 & 3.151 & 3.226 & 3.603 & 2.803 & 1.679 \\
          \hline
          \multicolumn{10}{c}{Multimodal Methods} \\
          \hline
          ViPC~\cite{zhang2021view}$^{*}$               & 3.308 & 1.760 & 4.558 & 3.183 & 2.476 & 2.867 & 4.481 & 4.990 & 2.197  \\
          \hline
          CSDN~\cite{zhu2023csdn}$^{*}$                 & 2.570 & 1.251 & 3.670 & 2.977 & 2.835 & 2.554 & 3.240 & 2.575 & 1.742 \\
          \hline
          XMFNet~\cite{aiello2022cross}            & 1.443 & 0.572 & 1.980 & 1.754 & 1.403 & 1.810 & 1.702 & 1.386 & 0.945 \\
          \hline
          Ours                                     & \textbf{1.159} & \textbf{0.539} & \textbf{1.793} & \textbf{1.599} & \textbf{1.193} & \textbf{0.729} & \textbf{1.512} & \textbf{1.151} & \textbf{0.756} \\
          \hline
    \end{tabular}
    \end{table*}
    \begin{table*}[!t]
    \tiny
    \renewcommand\arraystretch{1.2}
        \centering
        \caption{Quantitative comparisons to state-of-the-art methods on known eight categories of ShapeNet-ViPC dataset by using Mean F-Score @ 0.001 with 2048 points. The best is highlighted in bold. $*$ means the code is not available or uncompleted.}
        \label{tab: fscore of constrative experiments}
        \footnotesize
        \begin{tabular}{c|c|c|c|c|c|c|c|c|c}
        \hline
        \multirow{2}{*}{Methods} & 
        \multicolumn{9}{c}{F-Score@0.001 (higher is better)} \cr\cline{2-10}
                                                    & Mean & Airplane & Cabinet & Car & Chair & Lamp & Sofa & table & Watercraft \cr
        \hline
        \hline
              \multicolumn{10}{c}{Unimodal Methods} \\
              \hline
              AtlasNet~\cite{groueix2018papier}     & 0.410 & 0.509 & 0.304 & 0.379 & 0.326 & 0.426 & 0.318 & 0.469 & 0.551 \\
              \hline
              FoldingNet~\cite{yang2018foldingnet}  & 0.331 & 0.432 & 0.237 & 0.300 & 0.204 & 0.360 & 0.249 & 0.351 & 0.518 \\
              \hline
              PCN~\cite{yuan2018pcn}                & 0.407 & 0.578 & 0.27 & 0.331 & 0.323 & 0.456 & 0.293 & 0.431 & 0.577 \\
              \hline
              TopNet~\cite{tchapmi2019topnet}       & 0.467 & 0.593 & 0.358 & 0.405 & 0.388 & 0.491 & 0.361 & 0.528 & 0.615 \\
              \hline
              ECG~\cite{pan2020ecg}                 & 0.704 & 0.880 & 0.542 & 0.713 & 0.671 & 0.689 & 0.534 & 0.792 & 0.810 \\
              \hline
              VRC-Net~\cite{pan2021variational}     & 0.764 & 0.902 & 0.621 & 0.753 & 0.722 & 0.823 & 0.654 & 0.810 & 0.832 \\
              \hline
              PF-Net~\cite{huang2020pf}             & 0.551 & 0.718 & 0.399 & 0.453 & 0.489 & 0.559 & 0.409 & 0.614 & 0.656 \\
              \hline
              MSN~\cite{liu2020morphing}            & 0.578 & 0.798 & 0.378 & 0.380 & 0.562 & 0.652 & 0.410 & 0.615 & 0.708 \\
              \hline
              GRNet~\cite{xie2020grnet}             & 0.601 & 0.767 & 0.426 & 0.446 & 0.575 & 0.694 & 0.450 & 0.639 & 0.704 \\
              \hline 
              PoinTr~\cite{yu2021pointr}            & 0.683 & 0.842 & 0.516 & 0.545 & 0.662 & 0.742 & 0.547 & 0.723 & 0.780 \\
              \hline
              PointAttN~\cite{wang2022pointattn}    & 0.662 & 0.841 & 0.483 & 0.515 & 0.638 & 0.729 & 0.512 & 0.699 & 0.774 \\
              \hline
              SDT~\cite{zhang2022point}             & 0.473 & 0.636 & 0.291 & 0.363 & 0.398 & 0.442 & 0.307 & 0.574 & 0.602 \\
              \hline
              Seedformer~\cite{zhou2022seedformer}  & 0.688 & 0.835 & 0.551 & 0.544 & 0.668 & 0.777 & 0.555 & 0.716 & 0.786 \\
              \hline
              \multicolumn{10}{c}{Multimodal Methods} \\
              \hline
              ViPC~\cite{zhang2021view}$^{*}$             & 0.591 & 0.803 & 0.451 & 0.512 & 0.529 & 0.706 & 0.434 & 0.594 & 0.730  \\
              \hline
              CSDN~\cite{zhu2023csdn}$^{*}$               & 0.695 & 0.862 & 0.548 & 0.560 & 0.669 & 0.761 & 0.557 & 0.729 & 0.782 \\
              \hline
              XMFNet~\cite{aiello2022cross}         & 0.796 & 0.961 & 0.662 & 0.691 & 0.809 & 0.792 & 0.723 & 0.830 & 0.901\\
              \hline
              Ours                                  & \textbf{0.842} & \textbf{0.971} & \textbf{0.705} & \textbf{0.729} & \textbf{0.844} & \textbf{0.924} & \textbf{0.767} & \textbf{0.862} & \textbf{0.935} \\
        \hline
        \end{tabular}
    \end{table*}

\begin{figure}[h]
  \centering
  \includegraphics[width=\linewidth]{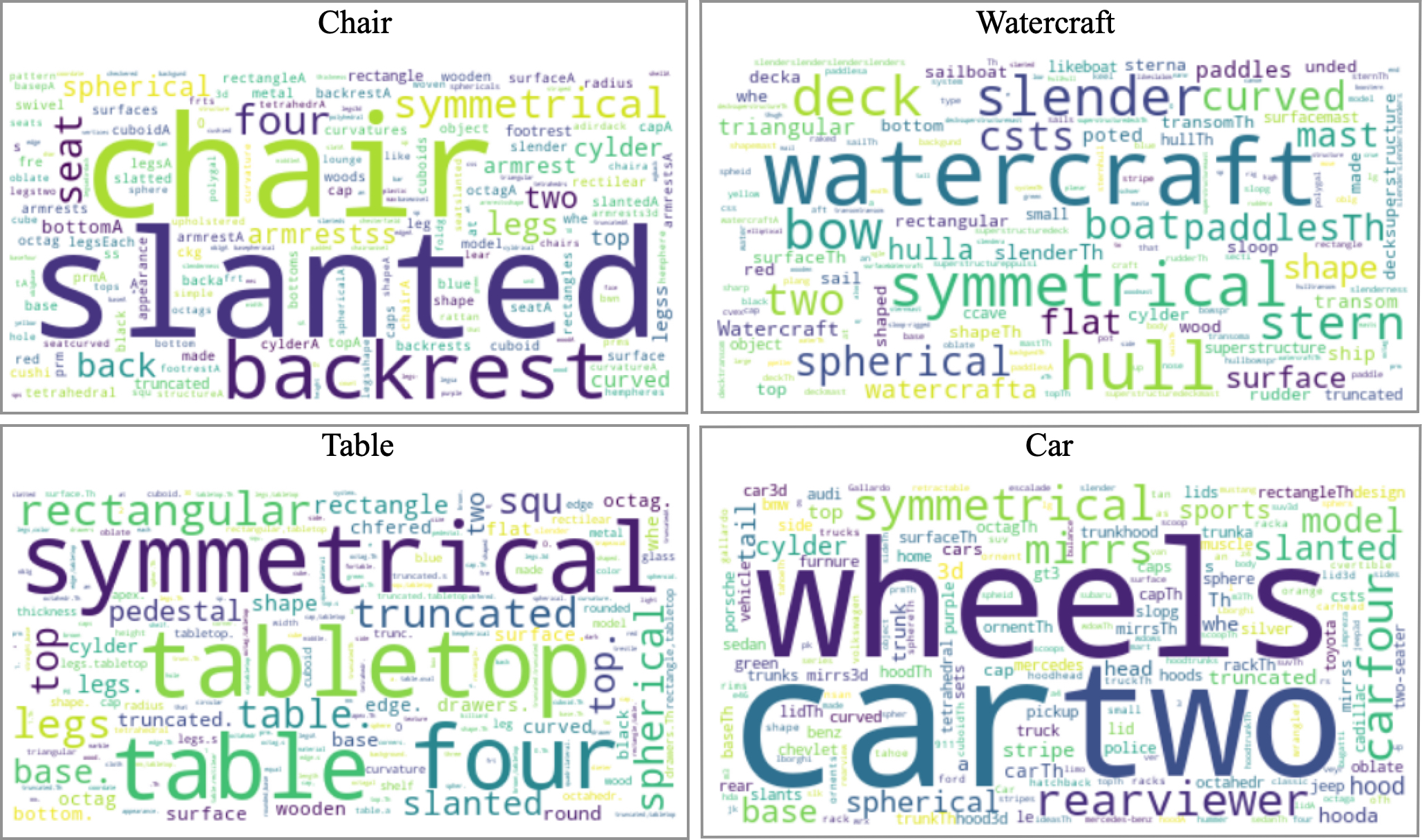}
\caption{Word cloud of ViPC-Text. Here, we show the word clouds of the most popular words found in the four subsets of the ViPC-Text dataset (including chair, table, car, and watercraft categories) by their frequency.}
  \label{fig:vipc_text}
\end{figure}

\begin{figure*}[h] 
  \centering
  \includegraphics[width=\textwidth]{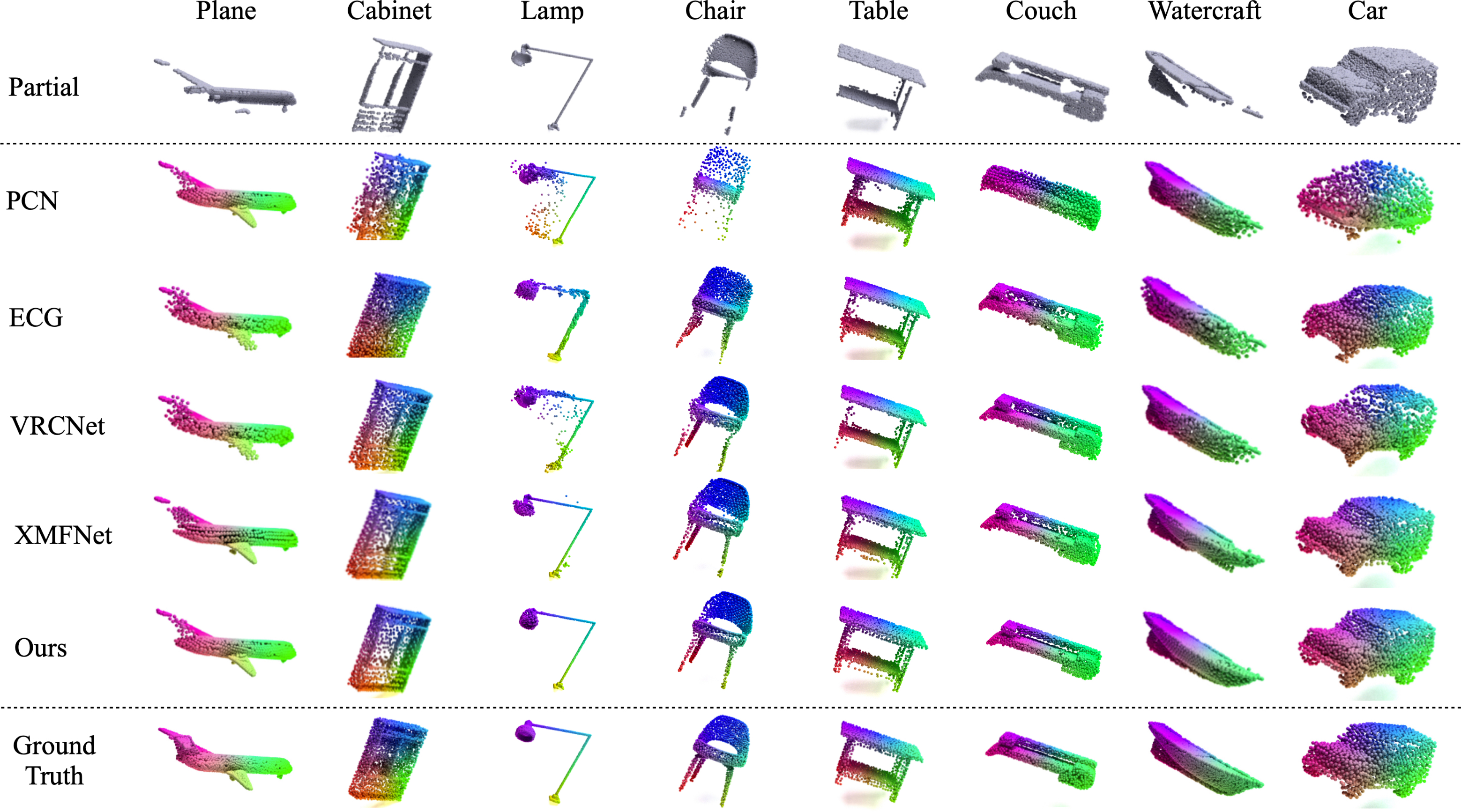}
  \caption{Qualitative comparisons between different methods including both the unimodal methods (PCN~\cite{yuan2018pcn}, ECG~\cite{pan2020ecg}, VRCNet~\cite{pan2021variational}) and multimodal methods (XMFNet~\cite{aiello2022cross} and our method).}
  \label{fig:contrastive_experiments}
\end{figure*}

\section{Experiments}\label{sec: Experiments and Results}
To verify the effectiveness of our algorithm and the constructed fine-grained geometric corpus, we review all the experiments and conduct exhaustive analysis in this section. First of all, we introduce the dataset used in our experiment in Sec.~\ref{subsec: Datasets} and provide more details about the critical settings of our experiment in Sec.~\ref{subsec: Implementation Details}. Then, see Sec.~\ref{subsec: comparisions on shapenet-vipc}, we show the comparisons of various point cloud completion methods on known and novel categories. Finally, the ablation experiments are also explained in Sec.~\ref{subsec: ablation} to further illustrate the effectiveness of our algorithm.
   
    \subsection{Datasets}\label{subsec: Datasets}
        \subsubsection{ShapeNet-ViPC}\label{ShapeNet-ViPC}
        In our experiments, we trained and tested the proposed method on a widely used ShapeNet-ViPC dataset~\cite{zhang2021view}, which contains a total of 13 category 3D shapes. Wherein 8 categories of 3D shapes, namely plane, lamp, cabinet, chair, table, sofa, boat, and car are used for training and evaluation. The remaining 5 categories of 3D shapes including bench, monitor, speaker, firearm and cellphone are only used for final zero-shot evaluation. The split of the dataset for training and evaluation is the same as XMFNet~\cite{aiello2022cross}.
    
    \subsubsection{ViPC-Text}\label{text corpus}
    To explore the effectiveness of rich text descriptions in our network, we build a new text corpus called ViPC-Text based on the ShapeNet-ViPC~\cite{zhang2021view}. It also contains 38,328 triples consisting of the text descriptions, 3D object and a set of rendered images from 24 view angles. Some example triples are visualized in the Fig.~\ref{fig:vipc_text}. It can be seen that the text descriptions from our corpus accurately describes the rich fine-grained geometric details of the 3D shapes. Besides, the length of the text descriptions in our corpus are range in 50-58, which is relatively concise. The concise text can be directly used as the input of the CLIP model, which can only accept the input of the token length less than 77. It should be emphasized that our text description is relatively concise, but our corpus contains a variety of vocabulary related to geometric descriptions to accurately describe 3D shapes. The word cloud of the  chair category in the corpus is shown in Fig.~\ref{fig:vipc_text}, our corpus contains a broad vocabulary related to number of shape components  and the geometric appearance of 3D shapes.
    
    \subsection{Implementation Details}\label{subsec: Implementation Details}
    \noindent\textbf{Corpus Generation.} The corpus generation algorithm is implemented on a single NVIDIA RTX A6000. In fact, the algorithm takes about 26G GPU memory and cost an average of 1 to 4 seconds to generate one item of text description from an image.

    \noindent\textbf{Network Training.} During training, we use the same training setting as XMFNet~\cite{aiello2022cross}. Our method is implemented using PyTorch framework~\cite{paszke2019pytorch} and all our experiments are performed on a 4 GPU NVIDIA A800-SXM4-80G cluster with 320G GPU memory, occupying about 160G cluster's GPU memory when running in parallel. We use Adam optimizer~\cite{kingma2014adam} with $\beta_{1}$ = 0.9 and $\beta_{2}$ = 0.999. The model is trained for 400 epochs with a batch size of 560 and an initial learning rate of 0.00209.

    \begin{figure*}[htb]
    \centering
    \includegraphics[width=\textwidth]{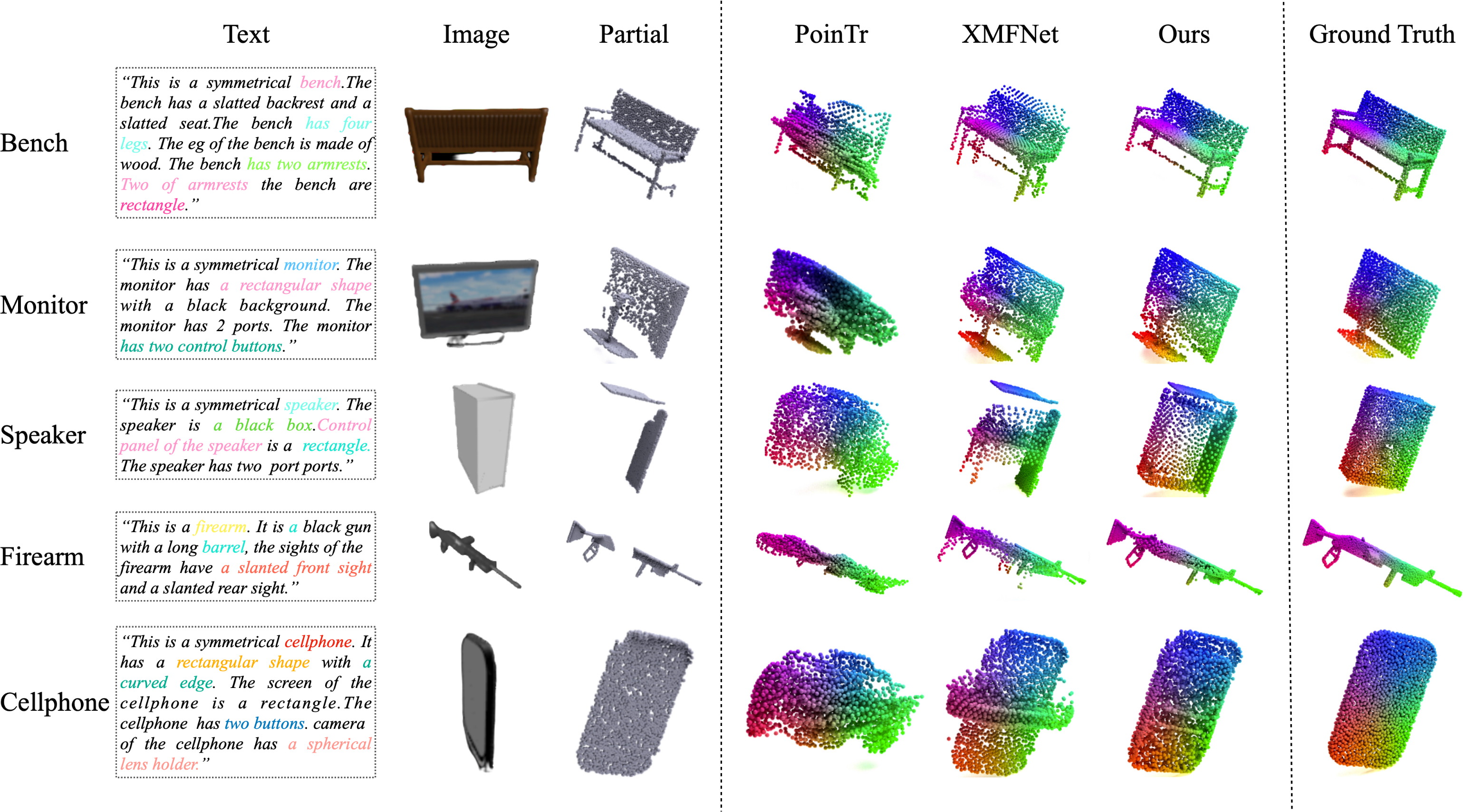}
        \caption{Visual comparisons of recent point cloud completion methods~\cite{yu2021pointr, aiello2022cross} and ours on unseen categories of ShapeNet-ViPC~\cite{zhang2021view}. Our method produces the most complete and detailed structures compared to its competitors.}
    \label{fig: zero-shot}
    \end{figure*}
 
    \subsection{Comparisons on ShapeNet-ViPC Dataset}\label{subsec: comparisions on shapenet-vipc}
    \subsubsection{Comparisons on Known Categories}\label{comparisons on known categories}

    In this section, we compare our method with other unimodal and multimodal methods. First, to quantitatively evaluate the performance of multiple methods, we use CD and F-score as metrics for the reconstruction quality on the ShapeNet-ViPC dataset. The evaluation metrics are similar to XMFNet\cite{aiello2022cross}. Tab.~\ref{tab: cd of constrative experiments} and Tab.~\ref{tab: fscore of constrative experiments} report the quantitative results of multiple methods and it can be clearly seen that our method has great advantages over both unimodal and multimodal methods. Especially compared with the state-of-the-art multimodal fusion method XMFNet, our results show a significant improvement. Then, the qualitative comparisons are shown in Fig.~\ref{fig:contrastive_experiments}. Similar to XMFNet~\cite{aiello2022cross}, we compare our method with several unimodal completion methods including PCN~\cite{yuan2018pcn}, ECG~\cite{pan2020ecg}, VRCNet~\cite{pan2021variational}, while for the multimodal method, we only compare with XMFNet~\cite{aiello2022cross} for the reason that the code of ViPC is incomplete and the code of CSDN is not available. First of all, compared with other methods, our method can produce complete shapes with less noise points. Then, according to the visualization results, the proposed network has a stronger ability to predict the semantic information of missing parts and the reconstruction quality of the missing parts are more accurate. As shown in Fig.~\ref{fig:contrastive_experiments}, our method can reconstruct a more complete aircraft tail and lampshade for the airplane and lamp models respectively. And for car model in Fig.~\ref{fig:contrastive_experiments},  our method can also reconstruct more accurate car tires. In addition, thanks to the introduction of rich text information in our work, we can also predict the geometric structure of shapes more accurately than other methods. The Fig.~\ref{fig:contrastive_experiments} clearly shows that the slender structure of a lamp, the complex structure of the chair back, the flat structure of the table pedal and the corner of the watercraft head can all be better reconstructed compared with other methods. Therefore, in summary, our method is capable of producing cleaner, more structured and more semantic completions than the other methods.

    \begin{figure*}[h]
      \centering
      \includegraphics[width=0.98\textwidth]{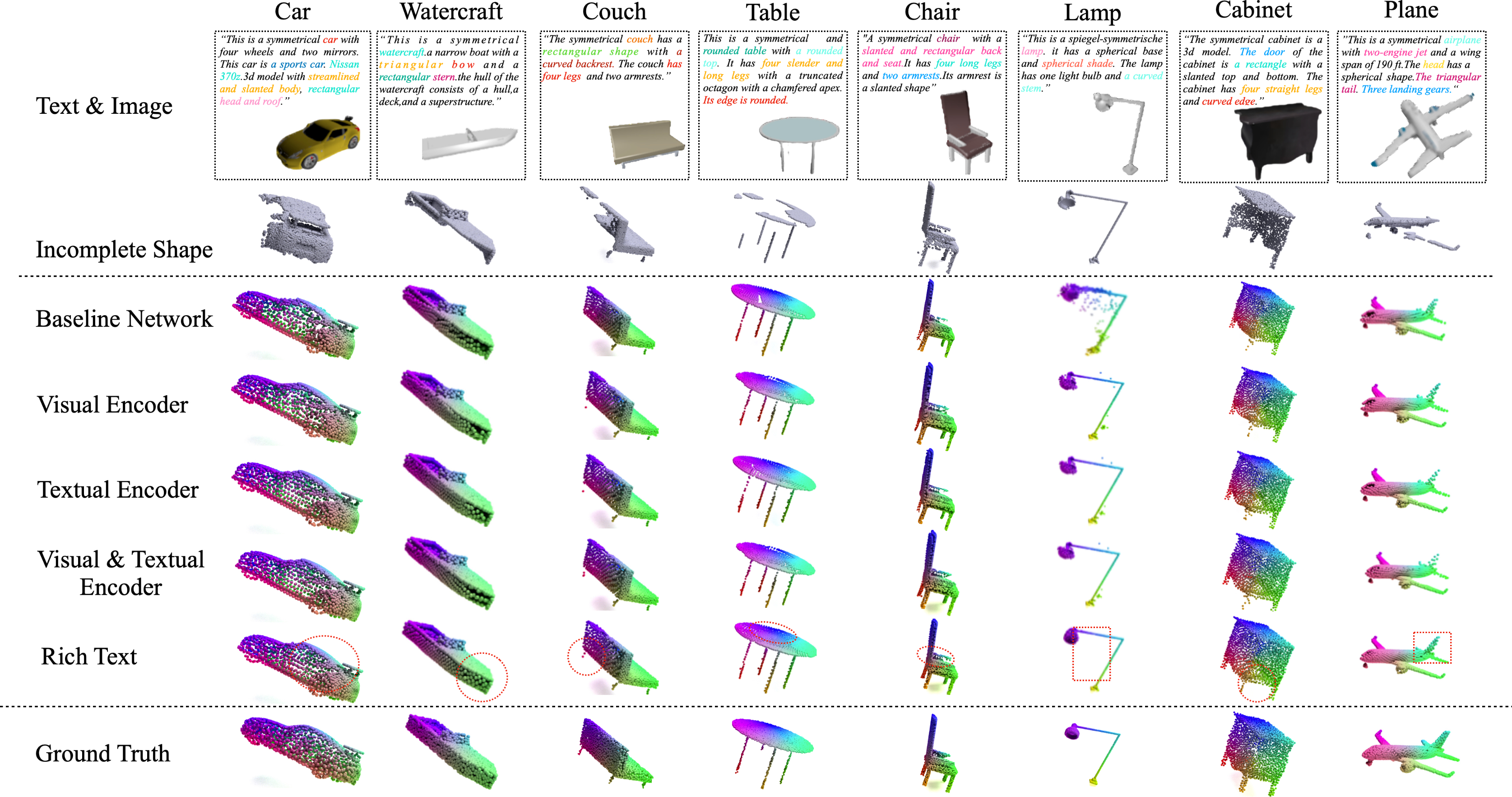}
    \caption{Visualized ablation study of eight known category objects on ShapeNet-ViPC dataset. The first row shows descriptions of the complex text; Rows 2-3 show the incomplete point clouds and the rendered image; Rows 4-9 show the visualization results under the use of different modal information; The last row gives the ground truth.}
      \label{fig: ablation_result}
    \end{figure*}

    \subsubsection{Comparisons on Novel Categories}\label{subsec: comparisons on novel categories}
    \begin{table}
        \renewcommand\arraystretch{1.2}
        \centering
        \caption{Quantitative comparisons to state-of-the-art methods on unknown five categories of ShapeNet-ViPC dataset by using Mean Chamfer Distance per point ($\times10^{-3}$) with 2048 points. The best is highlighted in bold.}
        \label{tab: cd of zero-shot experiments}
        \footnotesize
        \begin{tabular}{c|c|c|c|c|c}
        \hline
        \multirow{2}{*}{Methods} & \multicolumn{5}{c}{Mean CD per point (lower is better)} \cr\cline{2-6}
        & Mean & Bench & Monitor & Speaker & Phone \cr
        \hline
              \hline
              PF-Net~\cite{huang2020pf} & 5.011 & 3.684 & 5.304 & 7.663 & 3.392 \\
              \hline
              MSN~\cite{liu2020morphing} & 4.684 & 2.613 & 4.818 & 8.259 & 3.047 \\
              \hline
              GRNet~\cite{xie2020grnet} & 4.096 & 2.367 & 4.102 & 6.493 & 3.422 \\
              \hline
              PoinTr~\cite{yu2021pointr} & 3.755 & 1.976 & 4.084 & 5.913 & 3.049 \\
              \hline
              ViPC~\cite{zhang2021view} & 4.601 & 3.091 & 4.419 & 7.674 & 3.219\\
              \hline
              PointAttN~\cite{wang2022pointattn} & 3.674 & 2.135 & 3.741 & 5.973 & 2.848 \\
              \hline
              SDT~\cite{zhang2022point} & 6.001 & 4.096 & 6.222 & 9.499 & 4.189 \\
              \hline
              CSDN~\cite{zhu2023csdn} & 3.656 & 1.834 & 4.115 & 5.690 & 2.985 \\
              \hline
              XMFNet~\cite{aiello2022cross} & 3.259 & 1.512 & 3.668 & 5.417 & \textbf{2.439} \\
              \hline
              Ours & \textbf{3.098} & \textbf{1.217} & \textbf{3.043} & \textbf{5.311} & 2.819 \\
              \hline
        \end{tabular}
        \end{table}
        
    \begin{table}[!t]
        \renewcommand\arraystretch{1.2}
            \centering
            \caption{Quantitative comparisons to state-of-the-art methods on unknown five categories of ShapeNet-ViPC dataset by using Mean F-Score @ 0.001 with 2048 points. The best is highlighted in bold.}  
            \label{tab: f-score of zero-shot experiments}
            \footnotesize
            \begin{tabular}{c|c|c|c|c|c}
            \hline
            \multirow{2}{*}{Methods}& 
            \multicolumn{5}{c}{F-Score@0.001 (higher is better)} \cr\cline{2-6}
            & Mean & Bench & Monitor & Speaker & Phone \cr
            \hline
                  \hline
                  PF-Net~\cite{huang2020pf} & 0.468 & 0.584 & 0.433 & 0.319 & 0.534 \\
                  \hline
                  MSN~\cite{liu2020morphing} & 0.533 & 0.706 & 0.527 & 0.291 & 0.607 \\
                  \hline
                  GRNet~\cite{liu2020morphing} & 0.548 & 0.711 & 0.537 & 0.376 & 0.569 \\
                  \hline
                  PoinTr~\cite{yu2021pointr} & 0.619 & 0.797 & 0.599 & 0.454 & 0.627 \\
                  \hline
                  ViPC~\cite{zhang2021view} & 0.498 & 0.654 & 0.491 & 0.313 & 0.535 \\
                  \hline
                  PointAttN~\cite{wang2022pointattn} & 0.605 & 0.764 & 0.591 & 0.428 & 0.637 \\
                  \hline
                  SDT~\cite{zhang2022point} & 0.327 & 0.479 & 0.268 & 0.197 & 0.362 \\
                  \hline
                  CSDN~\cite{zhu2023csdn} & 0.631 & 0.798 & 0.598 & 0.485 & 0.644 \\
                  \hline
                  XMFNet~\cite{aiello2022cross} & 0.664 & 0.830 & 0.622 & 0.517 & \textbf{0.687} \\
                  \hline
                  Ours  & \textbf{0.687}  & \textbf{0.872} & \textbf{0.656} & \textbf{0.542} & 0.678 \\
            \hline
            \end{tabular}
        \end{table} 
        
    To demonstrate the generalization ability of the proposed method, we also show the quantitative and qualitative results of five unknown categories on the ShapeNet-ViPC dataset~\cite{zhang2021view}. Specifically, for all methods that need to be compared, we train the more general models by using the known 8 categories on the ShapeNet-ViPC dataset~\cite{zhang2021view}, then we evaluate these models on the 4 unknown categories (Actually the same as CSDN~\cite{zhu2023csdn}, we only tested four unknown categories of objects). Still using the same metrics, the quantitative results of CD and F-Score are reported in Tab.~\ref{tab: cd of zero-shot experiments} and Tab.~\ref{tab: f-score of zero-shot experiments}, respectively. Our approach still outperforms state-of-the-art unimodal methods such as PointAttN~\cite{wang2022pointattn} and PoinTr~\cite{yu2021pointr} as well as multimodal method XMFNet~\cite{aiello2022cross}. Besides, we also present a visual comparison between our method and  other methods including a Transformer-based method PoinTr~\cite{yu2021pointr} and a multimodal-based method XMFNet~\cite{aiello2022cross}, as shown in Fig.~\ref{fig: zero-shot}. We can observe that PoinTr~\cite{yu2021pointr} cannot well recover the missing shapes for previously unseen categories, although it achieves competitive performance on quantitative results. Then, only with the help of image auxiliary information, the completed shapes produced by XMFNet~\cite{aiello2022cross} suffer from poor quality. In contrast, our method can produce clear results with stronger structural details. This comparison further demonstrates that our method successfully exploits the complementary information provided by the images and text for point cloud completion.

    \subsection{Ablation Study}\label{subsec: ablation}
    \begin{table*}[!t]
        \tiny
        \renewcommand\arraystretch{1.2}
        \centering
        \caption{Ablation studies for the multimodal information used in our network on the ShapeNet-ViPC dataset. The Chamfer Distance is used as the metric for evaluating the gains from different multimodal information and different fusion strategies.}
        \label{tab: For CD ablation experiments}
        \scriptsize
        \begin{tabular}{c|c|c|c|c|c|c|c|c|c|c|c|c}
            \hline
            \multirow{2}{*}{\textbf{Ablation}} &
            \multirow{2}{*}{\tabincell{c}{Fusion\\ Stage 1}}&
            \multirow{2}{*}{\tabincell{c}{Fusion\\ Stage 2}}&
            \multicolumn{10}{c}{Mean CD per point $\times 10^{-3}$ (lower is better)}\\
            \cline{4-13} 
            &~ &~ &\textbf{Mean} & \textbf{Improv.(\%)} & Airplane & Cabinet & Car & Chair & Lamp & Sofa & table & Watercraft \\
            \hline
            \hline
            Baseline (XMFNet~\cite{aiello2022cross}) & -  & - & 1.443 & - & 0.572 & 1.980 & 1.754 & 1.403 & 1.810 & 1.702 & 1.386 & 0.945 \\
            \hline
            w/ Visual Encoder &\checkmark &\checkmark & 1.206 & 16.42 & 0.546 & 1.859 & 1.717 & 1.227 & 0.790 & 1.539 & 1.181 & 0.786 \\
            \hline
            w/ Text Encoder & \checkmark & \checkmark & 1.251 & 13.31 & 0.564 & 1.814 & 1.736 & 1.346 & 0.881 & 1.554 & 1.320 & 0.791 \\
            \hline
            \multirow{3}{*}{\tabincell{c}{w/ Visual Encoder\\ \& Text Encoder}}         
            & \checkmark &~ & 1.190 & 17.53 & 0.546 & 1.808 & 1.682 & 1.204 & 0.791 & 1.545 & 1.153 & 0.787 \\
            \cline{2-13} 
           & ~& \checkmark & 1.188 & 17.67 & 0.542 & 1.830 & 1.690 & 1.181 & 0.791 & 1.541 & 1.144 & 0.786 \\
            \cline{2-13} 
            & \checkmark & \checkmark & 1.181 & 18.16 & 0.552 & 1.808 & 1.634 & 1.214 & 0.740 & 1.538 & 1.176 & 0.788 \\
            \hline
            Final(w/ Rich Text) & \checkmark & \checkmark & 1.159 & 19.68 & 0.539 & 1.793 & 1.599 & 1.193 & 0.729 & 1.512 & 1.151  & 0.756 \\
            \hline
            \end{tabular}
        \end{table*}
    \begin{table*}[!t]
            \tiny
            \renewcommand\arraystretch{1.2}
            \centering
            \caption{Ablation studies for the multimodal information used in our network on the ShapeNet-ViPC dataset. The Mean F-Score @ 0.001 is used as the metric for evaluating the gains from different multimodal information and different fusion strategies.}
       
            \label{tab: For F1_Score ablation experiments}
            \scriptsize
            \begin{tabular}{c|c|c|c|c|c|c|c|c|c|c|c|c}
                \hline
                \multirow{2}{*}{\textbf{Ablation}} & 
                \multirow{2}{*}{\tabincell{c}{Fusion\\ Stage 1}}&
                \multirow{2}{*}{\tabincell{c}{Fusion\\ Stage 2}}&
                \multicolumn{10}{c}{F-Score@0.001 (higher is better)} \\
                \cline{4-13}
                &~&~ & \textbf{Mean} & \textbf{Improv.(\%)} & Airplane & Cabinet & Car & Chair & Lamp & Sofa & table & Watercraft \\
                \hline
                \hline
                Baseline (XMFNet~\cite{aiello2022cross}) & - & - & 0.796 & - & 0.961 & 0.662 & 0.691 & 0.809 & 0.792 & 0.723 & 0.830 & 0.901 \\
                \hline
                w/ Visual Encoder & \checkmark & \checkmark & 0.835 & 4.90 & 0.969 & 0.695 & 0.709 & 0.839 & 0.914 & 0.762  & 0.859 & 0.929 \\
                \hline
                w/ Text Encoder & \checkmark & \checkmark & 0.823 & 3.39 & 0.967 & 0.700 & 0.699 & 0.814 & 0.890 & 0.757 & 0.831 & 0.929 \\
                \hline
                \multirow{3}{*}{\tabincell{c}{w/ Visual Encoder\\ \& Text Encoder}}   
                & \checkmark & & 0.831 & 4.40 & 0.929 & 0.698 & 0.712 & 0.842 & 0.914 & 0.758 & 0.863 & 0.929 \\
                \cline{2-13} 
                & & \checkmark & 0.836 & 5.03 & 0.970 & 0.694 & 0.710 & 0.844 & 0.915 & 0.760 & 0.864 & 0.930 \\
                \cline{2-13}                       
                & \checkmark & \checkmark & 0.838 & 5.28 & 0.968 & 0.703 & 0.722 & 0.840 & 0.922 & 0.763 & 0.860 & 0.930 \\
                 \hline
                Final(w/ Rich Text) & \checkmark & \checkmark & 0.842 & 5.78 & 0.971 & 0.705 & 0.729 & 0.844 & 0.924 & 0.767 & 0.862& 0.935 \\ 
                \hline
            \end{tabular}
        \end{table*}

    \begin{figure*}[htb]
      \centering
      \includegraphics[width=\textwidth]{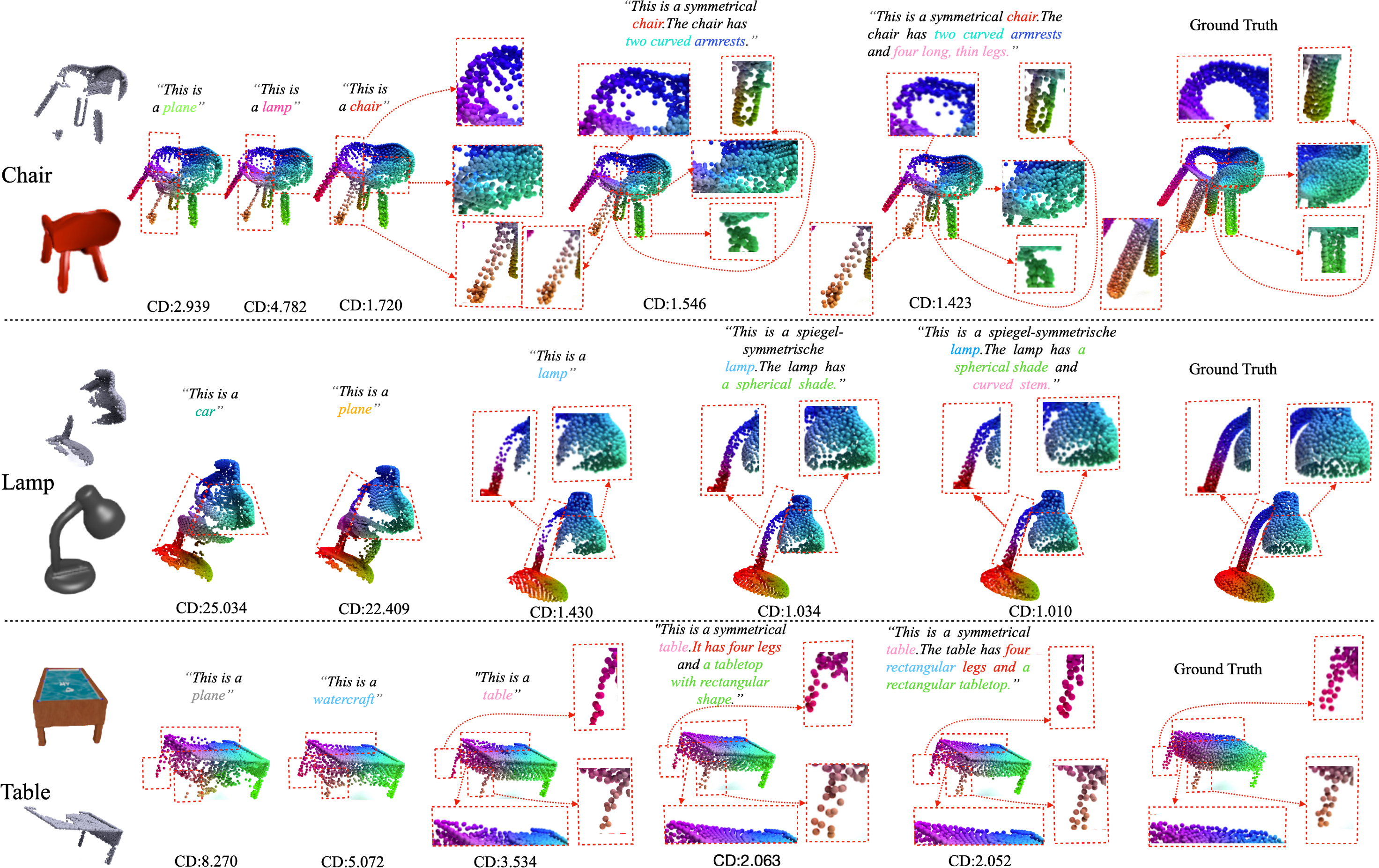}
    \caption{Ablation study of various textual information. From left to right, we gradually increase the text description information (wrong text information is given on the left, and accurate description details are given on the right), and the quality of the complete shapes is improved. In addition, the values below each example are the CD metric for evaluating the reconstruction quality between different text descriptions.}
      \label{fig: change_text}
    \end{figure*}

        \subsubsection{Performance of Multimodal Information}
        
        ~
        
        \noindent\textbf{The performance of $G_{i}$.} 
        As shown in the first and second rows of Tab.~\ref{tab: For CD ablation experiments} and Tab.~\ref{tab: For F1_Score ablation experiments}, we conduct an ablation study to access the effectiveness of the visual encoder of CLIP. Here, we only fuse the visual global features $G_{i}$ with the fine-grained features of incomplete point clouds. It should be noted that the input image used in CLIP is the same as the input in the baseline model. Compared with the baseline (XMFNet~\cite{aiello2022cross}), the introduction of the visual global feature in the CLIP can achieve 16.42\% and 4.90\% performance for the metrics of CD and F-Score on the known categories of ShapeNet-ViPC dataset~\cite{zhang2021view}. This experiment can verify that the visual module of the CLIP can deliver more generalized visual information compared with the fine-grained image features in the baseline, thereby achieving a significant improvement.
        
        ~

        \noindent\textbf{The performance of $G_{t}$.} 
        Compared to visual information, the semantic description text for point clouds is more readily available in practical applications. To verify the validity of the introduction of text information, we use a simple text to guide the process of the point cloud completion task. In fact, we only use the simple prompt “This is a [category]” as the input of our multimodal feature extraction module. The experimental results in the first and third rows of Tab.~\ref{tab: For CD ablation experiments} and Tab.~\ref{tab: For F1_Score ablation experiments} show that this simple global textual template is also useful for guiding the process of point cloud completion. As reported, the improvements of 13.31\% and 3.39\% can be accessed for the metrics of CD and F-Score, respectively. Therefore, the performances illustrate that just a simple text related to the object category can provide effective semantic information.
        
        ~

        \noindent\textbf{The performance of $G_{i}+G_{t}$.} 
        Considering the complementarity between the multimodal information, we also conduct an ablation study to verify the effectiveness of the textual and visual information fusion. As reported in the first and sixth rows of Tab.~\ref{tab: For CD ablation experiments} and Tab.~\ref{tab: For F1_Score ablation experiments}, when visual information and simple text information are both fused into our completion network, the improvements of 18.16\% and 5.28\% can be accessed for the metrics of CD and F-Score, respectively. This demonstrates that the complementarity among the three modal information of point cloud, image, and text description exists, and the quality of the reconstruction 3D shape can be better improved by using the three modal information at the same time. 
        In addition, we also verified the effectiveness of fine-grained geometric description text information, as shown in the last two rows of the Tab.~\ref{tab: For CD ablation experiments} and Tab.~\ref{tab: For F1_Score ablation experiments}, compared to the simple text, text description used in our ViPC-Text dataset can bring 1.86\% and 0.48\% improvements for the metrics of CD and F-Score. Ultimately, our method can access  19.68\% and 5.78\% improvement compared to the XMFNet~\cite{aiello2022cross}. Furthermore, as shown in the red dashed box of Fig.~\ref{fig: ablation_result}, our method with rich text descriptions has more advantages in structure and detail reconstruction. Our reconstruction results are sharper and the distribution of points is more uniform.
        
    \subsubsection{The Strategy of Multi-Stage Fusion}\label{subsec: Effect of Multi-Stage Fusion}
    As shown in the fourth to sixth rows of Tab.~\ref{tab: For CD ablation experiments} and Tab.~\ref{tab: For F1_Score ablation experiments}, we show an ablation study to explore  the strategies for multimodal feature fusion. Specifically, we compare three fusion strategies as follows:
    \begin{itemize}
        \item Multimodal feature fusion is done before fine-grained fusion on top of the baseline.
        \item Multimodal feature fusion is done after fine-grained fusion on top of the baseline.
        \item Multimodal feature fusion is performed in two stages, before and after fine-grained fusion on top of the baseline.
    \end{itemize}
    Through meticulous experimental comparisons, we found that the most effective fusion strategy for point cloud completion is to choose the third one described above, which adopts two-stage fusion and can better extract complementary information among multimodal information.

    \subsubsection{The performance of various text descriptions}
    In this section, we design various text descriptions to explore the impact of text richness on the final complete results. As shown in Fig.~\ref{fig: change_text}, we select three shapes from different categories to explore the effectiveness of the added textual information. First of all, we use simple text descriptions (e.g. “This is a [chair]”) to explore the influence of the textual information on the final complete shapes. Specifically, we compare visualization of the complete shapes predicted from correct and incorrect textual information. We found that misleading semantic text descriptions can severely affect model performance. The CD values below the instances are consistent with the visual results. In addition, we compared the impact of simple text descriptions and rich text descriptions on final complete shapes. The visualization shows that rich text information can provide the completion network with more additional geometric and semantic information than simple text descriptions. For example, for the lamp model in Fig.~\ref{fig: change_text}, the input of rich text description can improve the reconstruction quality of the cap and stem compared with the simple text description. For the chair model in Fig.~\ref{fig: change_text}, our network can predict the cleaner arms and legs when rich texts are used as the input. For the table model in Fig.~\ref{fig: change_text}, the rich text can improve reconstruction quality of tabletop and make the structure of the table legs closer to the rectangular structure. Consistently, the CD values under these visual examples can also reflect the effectiveness of rich text descriptions. Finally, we also tried to construct more precise hand-crafted descriptions to each example, as shown in the penultimate column of Fig.~\ref{fig: change_text}. It can be seen that a finer description can indeed improve the reconstruction accuracy of the incomplete shape. This also indicates that the text modality plays an important role in the task of point cloud completion. 
\label{secEXP}

\section{Conclusion}
In this paper, based on the CLIP model, we propose a novel multimodal fusion architecture called FTPNet for 3D point cloud completion. Different from existing multimodal methods, we introduce textual description into the completion architecture. Then, to further explore the function of textual information on the quality of 3d shape completion, we design an automatic fine-grained text generation method that can construct a multi-category fine-grained text corpus. Extensive experiments demonstrate that fine-grained text descriptions can dramatically improve our FTPNet's ability to understand the semantics and structure information of 3D shapes. However, our method still has some drawbacks where the model’s ability to understand the fine-grained text information is still poor. In the future, we will further explore the relationship between point cloud completion and fine-grained text, and build a fine-grained and controllable text-guided 3D point cloud completion framework.

\section*{Acknowledgments}
The authors would like to thank the High Performance Computing Center of Dalian Maritime University for providing the computing resources. Jun Zhou is supported by NSFC Fund (No.~62002040) and China Postdoctoral Science Foundation (No.~2021M690501). Xiuping Liu
is supported by NSFC Fund (No.~61976040). Mingjie Wang is supported by the Science Foundation of Zhejiang Sci-Tech University (No.~22062338-Y). Hongchen Tan is supported by National Natural Science Foundation of China (No.~62201020) and Beijing Postdoctoral Science Foundation (No.~2022-ZZ-069). 
\label{secCon}
\ifCLASSOPTIONcaptionsoff
  \newpage
\fi
\bibliographystyle{IEEEtran}
\bibliography{main_FTPNet}

\end{document}